\documentclass[10pt,twocolumn,letterpaper]{article}

\usepackage{cvpr}
\usepackage{times}
\usepackage{epsfig}
\usepackage{graphicx}
\usepackage{amsmath}
\usepackage{amssymb}
\usepackage{diagbox}
\usepackage{authblk}

\usepackage[dvipsnames]{xcolor}

\usepackage[linesnumbered,ruled,vlined]{algorithm2e}
\SetKwInput{KwInput}{Input}                
\SetKwInput{KwOutput}{Output}              

\DeclareMathOperator*{\argminA}{minimize} 

\usepackage[pagebackref=true,breaklinks=true,letterpaper=true,colorlinks,bookmarks=false]{hyperref}

\cvprfinalcopy 

\def\cvprPaperID{2704} 

\ifcvprfinal\fi
\begin{document}

\title{Self-Supervised Fast Adaptation for Denoising via Meta-Learning}

\author[1]{Seunghwan Lee}
\author[2]{Donghyeon Cho}
\author[3]{Jiwon Kim}
\author[1]{Tae Hyun Kim}
\affil[1]{Department of Computer Science, Hanyang University, Seoul, Korea

{\tt\small edltmd@hanyang.ac.kr, lliger9@gmail.com}}
\affil[2]{Department of Electronic Engineering, Chungnam National University, Daejeon, Korea

{\tt\small cdh12242@gmail.com}}
\affil[3]{SK T-Brain, Seoul, Korea

{\tt\small jk@sktbrain.com}}

\maketitle

\begin{abstract}
Under certain statistical assumptions of noise, recent self-supervised approaches for denoising have been introduced to learn network parameters without true clean images, and these methods can restore an image by exploiting information available from the given input (i.e., internal statistics) at test time. However, self-supervised methods are not yet combined with conventional supervised denoising methods which train the denoising networks with a large number of external training samples.
Thus, we propose a new denoising approach that can greatly outperform the state-of-the-art supervised denoising methods by adapting their network parameters to the given input through self-supervision without changing the networks architectures.
Moreover, we propose a meta-learning algorithm to enable quick adaptation of parameters to the specific input at test time.
We demonstrate that the proposed method can be easily employed with state-of-the-art denoising networks without additional parameters, and achieve state-of-the-art performance on numerous benchmark datasets.

\end{abstract}

\section{Introduction}
When a scene is captured by imaging devices, a desired clean image $\mathbf{X}$ is corrupted by noise $\mathbf{n}$. We usually assume that the noise $\mathbf{n}$ is an Additional White Gaussian Noise (AWGN), and the observed image $\mathbf{Y}$ can be expressed as $\mathbf{Y} = \mathbf{X} + \mathbf{n}$.
In particular, noise $\mathbf{n}$ increases in environments with high ISO, short exposure times, and low-light conditions.
Image denoising is a task that restores the clean image $\mathbf{X}$ by removing noise $\mathbf{n}$ from the noisy input $\mathbf{Y}$, and is a highly ill-posed problem. Thus, substantial literature concerning denoising problem has been introduced~\cite{Nuclear_denoise,wavelet_denoise,collaborative_denoise,nonlocal_sparse_denoise,centralized_sparse_denoise,dictionary_denoise,color_sparse_denoise,nonlocal_denoise}.

Recent deep learning technologies have been used not only to obtain an image prior model via discriminative learning but also to design feed-forward denoising networks that directly produce denoised outputs. These methods train networks for denoising by using pairs of input images and true clean images (Noise2Truth), and have performed well. However, 
Noise2Truth-based methods are limited in performance when the noise distribution of the test image is considerably different from the distribution of the training dataset,~\ie when domain misalignment occurs. To overcome these issues, researchers have proposed new training methods recently, such as Noise2Noise~\cite{Noise2noise}, Noise2Void~\cite{Noise2void}, and Noise2Self~\cite{Noise2self}, which allow to train the denoising networks without using the true clean images. These methods are based on statistical assumptions, such as zero-mean noise (i.e., $\mathbb{E}(\mathbf{n}) = 0$).

In this study, we improve the performance of existing Noise2Truth-based networks through a method that updates the network parameters adaptively using the information available from the given noisy input image. First, we start with a pre-trained network by the Noise2Truth technique to fully explore the large external database. Then, the network is fine-tuned using the Noise2Self method using the input test image during the inference phase. This approach not only solves the domain misalignment problem, but also improves the denoising performance by exploiting the self-similarity present in the input image. Self-similarity is a property that a large number of corresponding patches are existing within a single image (patch-recurrence), and it has been employed in numerous super-resolution tasks to enhance the restoration quality~\cite{glasner,zssr,selfex}.

We experimentally show that the adaptation via self-supervision during the inference stage can consistently increase denoising performance regardless of the deep learning architectures and target datasets. Furthermore, we adopt a meta-learning technique~\cite{reptile} to train the denoising networks to be quickly adapted to the specific input images at test time. 
Overall, our method obtains generalization based on Noise2Truth by using the large external training data while breaking the limit of previously achieved performance through adoption of the Noise2Self approaches. 

In this study, we present a new learning method which allows to train the denoising networks by supervision and self-supervision, and boosts the inference speed by training the network with a meta-learning algorithm.
To the best of our knowledge, this work is the first attempt to seriously explore meta-learning for the denoising task.
The contributions of this paper are summarized as follows:
\begin{itemize}
\item Conventional supervised denoising networks can be further improved by self-supervision during the test time. A two-phase approach, which utilizes the internal statistics of a natural image (self-supervision), is proposed to enhance restoration quality during the test time.
\item A meta-learning-based denoising algorithm, which facilitates the denoising network to quickly adapt parameters to the given test image, is introduced.
\item The proposed algorithm can be easily applied to many conventional denoising networks without changing the network architectures and improve performance by a large margin.
\end{itemize}

\section{Related Work}
In this section, we review numerous denoising methods with and without the use of true clean images for training. 

Image denoising is an actively studied area in image processing, and various denoising methods have been introduced, such as self-similarity-based methods~\cite{NLM,BM3D,SADCT}, sparse-representation-based methods~\cite{Sparse09,Sparse11}, and external database exploiting methods~\cite{Category_specific,External_category,Adaptive15,External15}. 
With the recent development of deep learning technologies, the denoising area also has been improved, and remarkable progress has been achieved in this field. 
Specifically, after Xie~\etal~\cite{Denoising12} adopted deep neural networks for denoising and inpainting tasks, numerous follow-up studies have been proposed~\cite{DnCNN,IrCNN,FFDNet,RDN,Nonlocal_color,Nonlocal_recurrent,Nonlocal_residual,CBDNet,RIDNet}.

Based on deep CNN, Zhang~\etal~\cite{DnCNN} proposed a deep neural network to learn a residual image with a skip connection between the input and output of the network, and accelerate training speed and enhance denoising performance.
Zhang~\etal~\cite{IrCNN} also proposed IRCNN to learn a Gaussian denoiser and this network can be combined with conventional model-based optimization methods to solve various image restoration problems such as denoising, super-resolution, and deblurring. 
Furthermore, Zhang~\etal~\cite{FFDNet} proposed a fast and efficient denoising network FFDNet, which takes cropped sub-images and a noise level map as inputs. In addition to being fast, FFDNet can handle locally varying and a wide range of noise levels. Zhang~\etal~\cite{RDN} introduced a very deep residual dense network (RDN) which is composed of multiple residual dense blocks. RDN achieves superior performance by exploiting all the hierarchical local and global features through densely connected convolutional layers and dense feature fusion. To incorporate long-range dependencies among pixels, Zhang~\etal~\cite{Nonlocal_residual} proposed a residual non-local attention network (RNAN), which consists of a trunk and (non-) local mask branches. In~\cite{Nonlocal_recurrent}, the non-local block was used with a recurrent mechanism to increase the receptive field of the denoising network. Recently, CBDNet~\cite{CBDNet} and RIDNet~\cite{RIDNet} were introduced to handle noise in real photographs where the noise level is unknown (blind denoising). CBDNet is a two-step approach that combines noise estimation and non-blind denoising tasks, whereas RIDNet is a single-stage method that employs feature attention.

After deep CNN was adopted to increase denoising performance, various research directions, such as residual learning for constructing deeper networks, non-local or hierarchical features for enlarging the receptive fields, and noise level estimation for real photographs, have been considered. However, such works remain limited to the cases in which networks are supervised by true clean images (Noise2Truth). Recently, several self-supervision-based studies have been conducted to leverage only noisy images for network training without true clean images. Lehtinen~\etal~\cite{Noise2noise} demonstrated that a denoising network can be trained without clean images. The network was trained with pairs of noisy patches (Noise2Noise) based on statistical reasoning that the expectation of randomly corrupted signal is close to the clean target data. Furthermore, to avoid constructing pairs of noisy images, Krull~\etal~\cite{Noise2void} proposed a Noise2Void method and introduced a blind-spot network. Specifically, only the center pixel of the input patch was considered in the loss function, and the network was trained to predict its center pixel without any true clean dataset. Similarly, Baston and Royer~\cite{Noise2self} introduced a Noise2Self method for training the network without knowing the ground truth data.

However, these self-supervision-based methods can not outperform supervised methods where the distribution of the input is identical to training sample distribution.
We use the supervised approach (i.e., Noise2Truth) during the training phase to achieve state-of-the-art performance by generalization, and use a self-supervised approach (e.g., Noise2Void, Noise2Self) on the test input image during the inference stage to further improve the performance by adaptation. To do so, we can employ the conventional meta-learning algorithms~\cite{maml,reptile,metasgd} with our denoising networks to enable quick adaption of the network parameters to the given input image.
In the end, our supervised network can be adapted to the input image during the inference phase based on the self-supervision with only few gradient update steps.

The proposed method can achieve state-of-the-art performance by exploring the large external datasets, and exploiting internal information available from the given input image, such as self-similarity as in~\cite{Noise2noise,Noise2self,Noise2void}. To the best of our knowledge, this work is the first attempt to apply the meta-learning to enable quick adaptation with self-supervision for blind and non-blind denoising tasks.

\section{Supervision vs. Self-Supervision}

Recent learning-based denoising works~\cite{Noise2noise,Noise2self,Noise2void} have attempted to remove noise and restore the clean image by self-supervision without relying on a large training dataset.
Natural images have similar patches that are redundant within a single image~\cite{glasner,selfex,zssr}; thus, we can estimate clean patches by using the corresponding but differently corrupted patches, assuming that the expected value of the added random noise is zero~\cite{Noise2self, Noise2void}.

In general, these self-supervised methods can effectively remove unseen noise (i.e., noise from an unknown distribution) by exploiting self-similarity with specially designed loss functions at test-time, whereas conventional supervised methods cannot handle unexpected and unseen noise which is not sampled from the trained distribution.
Conventional supervised denoising networks that learned using a large external dataset cannot exploit self-similarity at test-time due to the limited capacity of the network architectures (e.g., receptive field), and thus the performance is limited.
In contrast, self-supervision-based methods cannot outperform the conventional supervision-based methods when the noisy input image is sampled under a learned distribution, 
because self-supervised methods do not learn from a large external dataset.

Therefore, we aim to improve the performance of the conventional supervised methods by merging supervised and self-supervised methods to utilize large external datasets and exploit the given test image.
However, integration techniques have yet to be investigated actively.

We first simply combine the self-supervision method and the conventional supervised Gaussian denoising network by using the fully pre-trained parameters of the Gaussian denoiser as initial parameters of the self-supervised network. After initialization, the parameters of the self-supervised network are updated (fine-tuned) using the test input without knowing the ground truth version, as in~\cite{Noise2void}.
However, as shown in Table~\ref{table_comp_n2v}, this naive integration even degrades the performance of the supervised baseline model~\cite{DnCNN} when the test image is corrupted by noise with learned distribution (i.e., Gaussian noise).
Therefore, in this work, we present a novel denoiser that improves restoration performance by deriving the benefits of a large external training dataset and self-similarity from an input image. Such benefits are derived by integrating both supervised and self-supervised methods.

\begin{table}[]
\begin{tabular}{lllll}
\cline{1-3}
\multicolumn{1}{|l|}{} & \multicolumn{1}{l|}{\begin{tabular}[c]{@{}l@{}}Supervision~\cite{DnCNN} \end{tabular}} & \multicolumn{1}{l|}{\begin{tabular}[c]{@{}l@{}}Supervision~\cite{DnCNN} +\\  
Self-supervision~\cite{Noise2void}
\end{tabular}} &  &  \\ \cline{1-3}
\multicolumn{1}{|l|}{\begin{tabular}[c]{@{}c@{}}PSNR\\ $(\sigma=40)$\end{tabular}} & \multicolumn{1}{c|}{27.84}                                                                            & \multicolumn{1}{c|}{27.39} &  &  \\
\cline{1-3} & & &  & 
\end{tabular}
\caption{Gaussian denoising results with and without using self-supervision. The backbone network of the self-supervision based method N2V~\cite{Noise2void} is DnCNN~\cite{DnCNN}.
N2V is initialized with fully trained parameters then updated using the input image as in~\cite{Noise2void}.
Notably, naive integration degrades the performance of the baseline model (i.e., DnCNN).}
\label{table_comp_n2v}
\end{table}

\section{Proposed Method}
\subsection{Two-phase denoising approach}
\begin{figure*}[t]
\begin{center}
\includegraphics[width=1.0\linewidth]{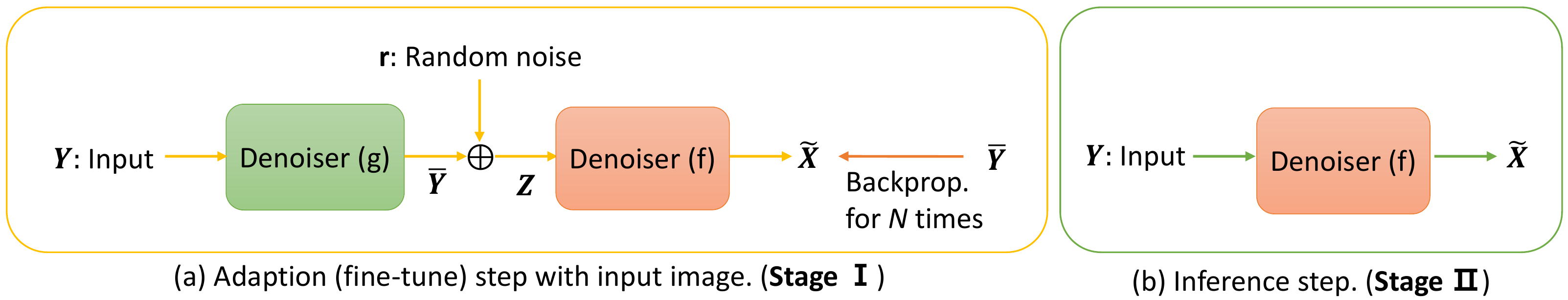}
\end{center}
\caption{Overall flow of the proposed method. Note that the denoiser $g$ is non-trainable while $f$ is trainable.}
\label{fig_overall}
\end{figure*}

\begin{figure}[t]
\begin{center}
\includegraphics[width=1.0\linewidth]{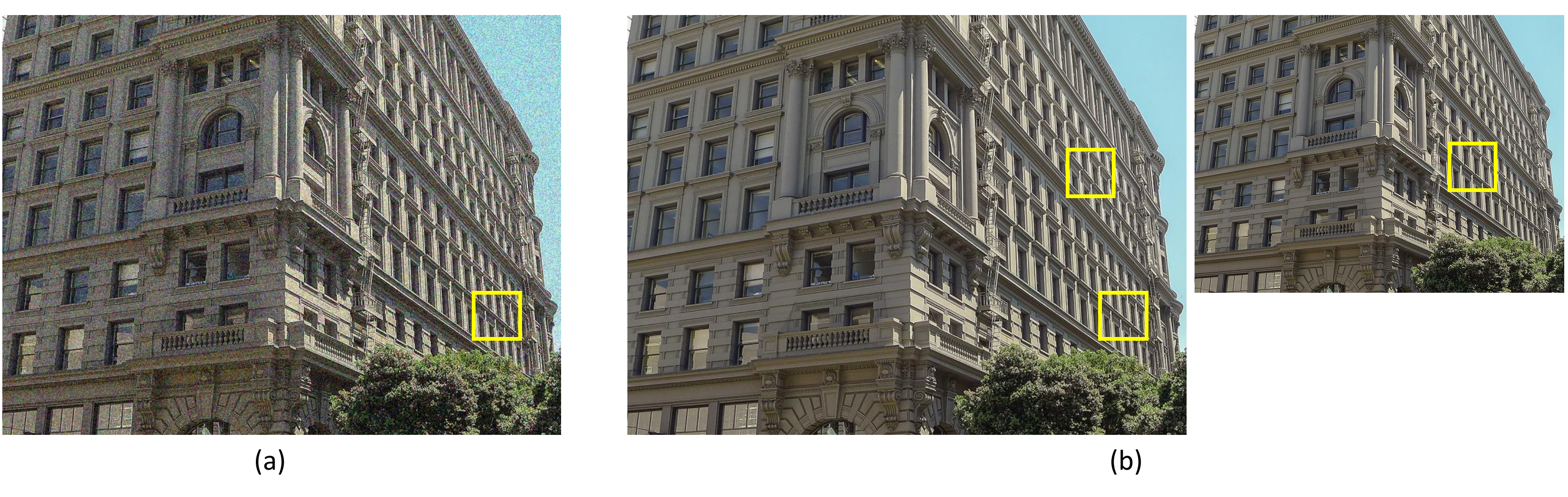}
\end{center}
\caption{(a) Noisy input image. (b) Denoised images at different image scales. Yellow patches are corresponding to each other. Clean (yellow) patches at different image scales in (b) can be used to remove the noise within the (yellow) patch in (a).}
\label{fig_self_sim}
\end{figure}

Many self-supervised methods~\cite{Noise2noise,Noise2void,Noise2self} assume that the input image is corrupted by independent and identically distributed (i.i.d) noise $\mathbf{n}$, and an optimal denoiser for the input can be estimated by minimizing the self-supervised loss function as follows:
\begin{equation}
\begin{split}
    Loss(\theta) = &\mathbb{E}\|f(\mathbf{Y}_i;\theta)-\mathbf{Y}_j\|^2 \\
    =&\mathbb{E}\|f(\mathbf{Y}_i;\theta)-\mathbf{X}_i\|^2 + \mathbb{E}\|\mathbf{Y}_j-\mathbf{X}_i\|^2,
\end{split}
\label{eq_n2n}
\end{equation}
where $\mathbf{Y}_i$ and $\mathbf{Y}_j$ are independently corrupted corresponding patches with i.i.d noise $\mathbf{n}$, $\mathbf{X}_i$ denotes their clean and ground-truth version, and $\mathbb{E}[\mathbf{Y}_i|\mathbf{X}_i]=\mathbf{X}_i$.
A mapping function $f$ is our denoiser and our goal is to estimate the parameters $\theta$.

Ideally, we can learn optimal parameters $\theta$ by minimizing the self-supervised loss with corresponding noisy patches~\cite{Noise2noise}.
Therefore, in the learning process, we should collect redundant and self-similar patches within the given image.
However, the number of corresponding patches is not infinitely many in practice; thus, minimizing $(\ref{eq_n2n})$ does not provide an optimal solution.
Moreover, finding corresponding patches within the given noisy image is a difficult and time-consuming task.
Therefore, recent self-supervised approaches slightly modify the loss function, and consider only the center pixel value as follows:
\begin{equation}
\begin{split}
    Loss(\theta) \approx \sum_i \|M_i(f(\mathbf{Y}_i;\theta))-M_{i'}(\mathbf{Y}_i)\|^2,
\end{split}    
\label{eq_n2v}
\end{equation}
where $M_i$ extracts a single pixel value at the center location of patch $\mathbf{Y}_i$, and $M_{i'}$ randomly takes a pixel value from the surrounding area of the center pixel within the same patch $\mathbf{Y}_i$, assuming that neighboring pixel values (e.g., color) are similar.
Therefore, the self-similarity exploitation ability is considerably limited because we can generate only 256 samples per patch (the number of samples can be used to calculate the expected value for $\mathbf{Y}_i$) because surrounding pixel values should be in between [0, 255] in gray-scale, and thus increase the variance of the estimator.
Moreover, self-supervised methods take much time in training because they compare only a single pixel value in the loss function while taking a large patch as input.

To alleviate this problem, we present a novel solution in this study. Specifically, we can reduce the amount of noise in the patch $\mathbf{Y}_i$ by using an arbitrary denoiser $g$, and we propose to minimize a new self-supervision loss as follows:
\begin{equation}
    Loss(\theta) = \mathbb{E}\|f(\mathbf{Y}_i;\theta)-\bar{\mathbf{Y}}_j\|^2,
\label{eq_loss_proposed1}
\end{equation}
where $\bar{\mathbf{Y}}_j$ denotes the denoised version of $\mathbf{Y}_j$ by the denoiser $g$. Note that $\mathbf{Y}_i$, and $\mathbf{Y}_j$ are corresponding.


If we assume that 
$\bar{\mathbf{Y}}_j = \mathbf{X}_i + \mathbf{n'}$
where the remaining (residual) noise $\mathbf{n'}$ is still i.i.d, then our denoiser $f$ can learn better parameters compared with those obtained by minimizing (\ref{eq_n2v}) because the noise level of $\bar{\mathbf{Y}}_j$ is lower than that of $\mathbf{Y}_j$ (i.e., $Var(\mathbf{n'}) < Var(\mathbf{n})$). 
In addition, we can generate a new noisy signal $\mathbf{Z}_j = \bar{\mathbf{Y}}_j + \mathbf{r}$ by adding some random noise $\mathbf{r}$ into the denoised patch $\bar{\mathbf{Y}_j}$,
and our loss function with respect to $\theta$ can be reformulated as
\begin{equation}
\begin{split}
    Loss(\theta) &= \mathbb{E}\|f(\mathbf{Z}_j;\theta)-\bar{\mathbf{Y}}_j\|^2\\
    &= \mathbb{E}\|f(\mathbf{Z}_i;\theta)-\bar{\mathbf{Y}}_i\|^2,
    \end{split}
    \label{eq_loss_proposed2}
\end{equation}
when the distribution of $\mathbf{Z}_j$ is identical to the distribution of $\mathbf{Y}_j$.
Therefore,  we can obtain the optimally denoised version of the $\mathbf{Y}_j$ by minimizing the proposed loss function, and it also becomes the clean version of $\mathbf{Y}_i$ because they are corresponding.
We no longer need to find corresponding patches from the given test image in (\ref{eq_loss_proposed2}), 
and we can compare patch by patch in the proposed loss function in contrast to the previous self-supervised works that calculate the loss pixel by pixel.

To be specific, if we generate $N$ noisy patches $\{\mathbf{Z}_i\}$ for the patch $\mathbf{Y}_i$, 
we can obtain a total of $MN$ self-similar patches when $M$ corresponding patches exist within the given image.
Then, the denoised patch $\tilde{\mathbf{X}}_i$ is given by,
\begin{equation}
    \tilde{\mathbf{X}}_i = \frac{1}{M} \sum_{i=1}^M (\frac{1}{N}\sum_{n=1}^N{\mathbf{Z}_{i}^{n}}),
\end{equation}
where $n$ denotes the index of the realized sample $\mathbf{Z}_i$.
When $N$ approaches infinity, $\tilde{\mathbf{X}}_i$ can be approximated as:
\begin{equation}
\begin{split}
    \tilde{\mathbf{X}}_i \approx \frac{1}{M} \sum_{j=1}^M \bar{\mathbf{Y}}_i=\frac{1}{M} \sum_{i=1}^M (\mathbf{X}_i+\mathbf{r}),
\end{split}
\label{eq_why_two_phase}
\end{equation}
and the denoised patch becomes the average value of $M$ corresponding patches denoised by $g$.

Ideally, we can generate a maximum of $N=256^{H \times W}$ samples from an $H \times W$ patch $\bar{\mathbf{Y}}_i$, and the variance of our estimator can be reduced by a factor of $M$ as $N \rightarrow \infty $.
However, the previous self-supervised methods can generate only a limited number of samples ($N$=256) per patch; thus, the variances of the estimators in \cite{Noise2self,Noise2void} are higher than our proposed estimator.
We can train the network $f$ with pairs of images ($\mathbf{Z}_i, \bar{\mathbf{Y}}_i$). Note that we can generate many $\mathbf{Z}_i$ correspond to $\bar{\mathbf{Y}_i}$, and thus the training procedure becomes super-efficient.
Moreover, in our experiments, the proposed loss function remains valid with images at different scales because self-similar patches are existing across scales~\cite{zssr,glasner}, as shown in Fig.~\ref{fig_self_sim}. This property allows the use of self-similarity in a larger space and can increase the number of corresponding patches $M$ within the given dataset.

Based on the proposed loss function in (\ref{eq_loss_proposed2}),
we present a new denoising network that can integrate the state-of-the-art supervised and self-supervised methods into a single network to utilize the power of deep learning with a large external database and internal statistics.
The sketch of the proposed two-phase denoising approach is illustrated in Fig.~\ref{fig_overall}.

\subsection{Fast adaptation via meta-learning}
\begin{figure}[h]
\begin{center}
\includegraphics[width=1.0\linewidth]{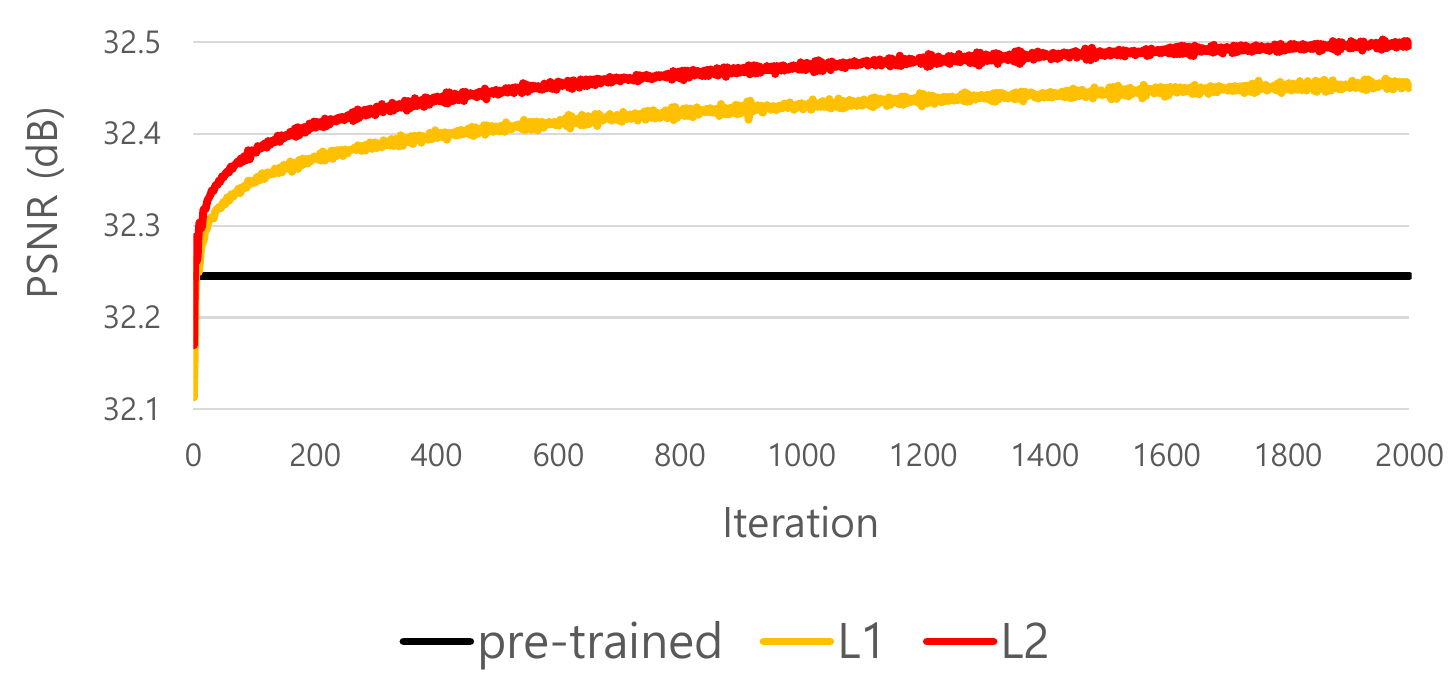}
\end{center}
\caption{Fully pre-trained DnCNN~\cite{DnCNN} on the DIV2K training set is given.
For each image $\mathbf{X}$ in the DIV2K test set, we generate 2000 train samples $\{\mathbf{Z}\}$, and minimize the proposed loss function in (\ref{eq_loss_proposed2}) at test time. The Average PSNR value goes up as iteration number increases. We also show the result by different metrics (i.e., $L1$ norm) which is also widely used in many recent works.}
\label{fig_finetune}
\end{figure}

\begin{figure*}[t]
\begin{center}
\includegraphics[width=0.9\linewidth]{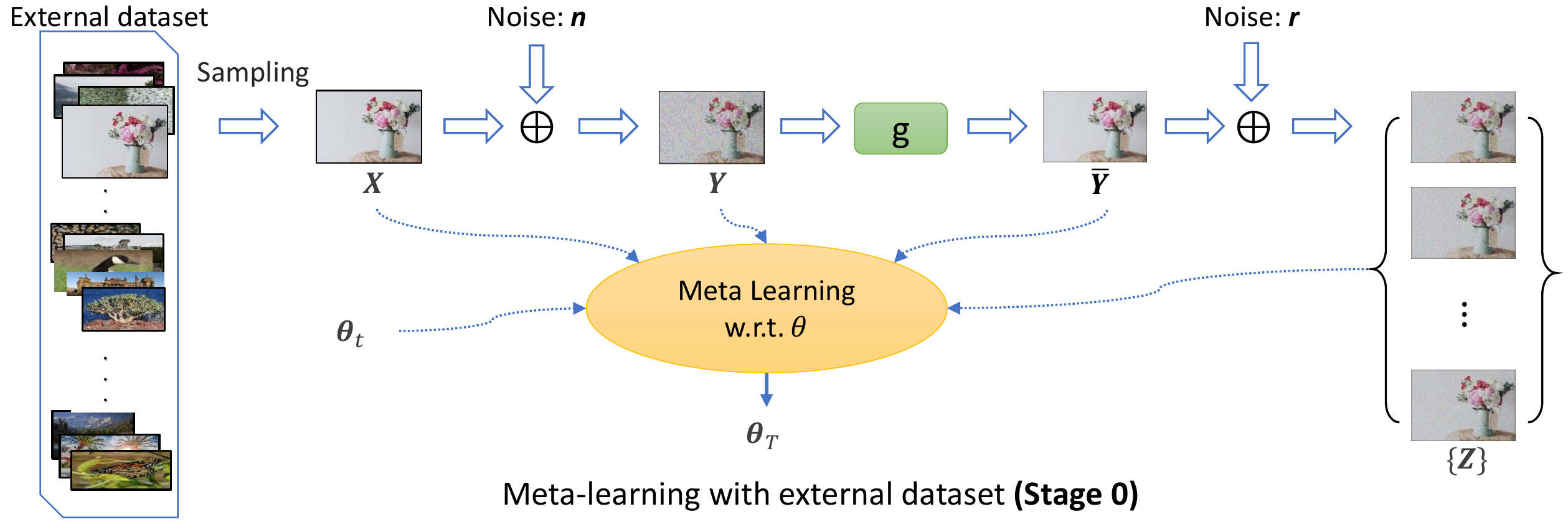}
\end{center}
\caption{Parameter initialization via meta-earning process with external large training dataset.}
\label{fig_meta_learn}
\end{figure*}

We can further update the parameters of fully trained denoising networks during the testing phase by minimizing the proposed loss function with the test input.
Fig.~\ref{fig_finetune} shows denoising performance of fully trained DnCNN~\cite{DnCNN} in terms of PSNR while updating the network parameters through the minimization of ($\ref{eq_loss_proposed2}$) using the DIV2K 10 validation set.
For the experiment, we use Gaussian noise for $\mathbf{n}$ and $\mathbf{r}$ ($\sigma$ = 20), and the fully pre-trained DnCNN on the DIV2K training set is used as $g$ and initial $f$.
According to the steps shown in Fig.~\ref{fig_overall}(a), we update the network $f$ with $\bar{\mathbf{Y}}$ and differently corrupted $\mathbf{Z}$ for 2000 iterations (i.e., $N=2000$ and mini-batch size = 1), and denoising performance improves as the update (fine-tune) procedure progresses, as shown in Fig.~\ref{fig_finetune}.
Notably, the PSNR value at iteration 0 denotes the performance of the initial $f$ (i.e., black solid line). Although PSNR drops for the first few iterations, we can elevate the performance of $f$ up to approximately 0.25dB through 2000 updates without using the ground truth image $\mathbf{X}$.

However, as we use the full-resolution image during the update procedure in Fig.~\ref{fig_overall}(a), it takes much time during the testing phase.
Therefore, we propose a fast update algorithm that allows a quick adaptation of the network parameters during the testing phase by embedding the recent meta-learning algorithms~\cite{maml,reptile,metasgd} into our two-phase denoising algorithm. 

Meta-learning algorithms can be used to find initial parameters of the network in the training stage, which facilitate fast adaptation at test time.
In general, meta-learning algorithms require ground-truth training samples for parameter adaptation at test time, but only a single noisy image is available in our denoising task. Thus, the use of the meta-learning scheme is restricted.
However, as shown in Fig.~\ref{fig_finetune}, we have shown that we can train the network $f$ in an unsupervised manner using $\bar{\mathbf{Y}}$ and a large number of $\{\mathbf{Z}\}$. 
Thus, we can adopt the meta-learning algorithms by using the training samples composed of $\bar{\mathbf{Y}}$ and $\{\mathbf{Z}\}$ to efficiently adapt our parameters at test time.
The overall flow of the proposed meta-learning process to initialize the parameters of $f$ for test-time adaptation with the external dataset is illustrated in Fig.~\ref{fig_meta_learn}.
Then, the meta-learned network parameters $\theta_T$ can be used as the initial parameters of $f$ for the test-time updates in the two-phase denoising algorithm. 

Specifically, our meta-learning integrated denoising algorithm is not limited to any specific meta-learning algorithm, and recent methods, such as MAML~\cite{maml}, Reptile~\cite{reptile}, and Meta-SGD~\cite{metasgd}, which aim for fast adaptation can be used. In our experiments, Reptile\cite{reptile} from OpenAI shows consistently better results compared with MAML~\cite{maml}. Thus, we provide the detailed steps of our meta-learning algorithm with Reptile~\cite{reptile} in Algorithm~\ref{algorithm_reptile}, and the inference algorithm during test time is given in Algorithm~\ref{algorithm_reptile_inference}. 
We believe Reptile outperforms MAML in our task, because our task requires relatively numerous iterations (updates) at test time.
Notably, our denoiser in Algorithm~\ref{algorithm_reptile_inference} can solve blind (unknown noise level) and non-blind (known noise level) denoising tasks without changing the training scheme in Algorithm~\ref{algorithm_reptile}. We only need to determine the noise level of $\mathbf{r}$ as a random or fixed value during test-time adaptation depending on the given task.

\begin{algorithm}[h]
\DontPrintSemicolon
  \KwInput{Fully pre-trained params.: $\theta_0$}
  \KwOutput{$\theta_T$}
  \KwData{Clean images $\{\mathbf{X}\}\newline$}

  \For{t = 1 to T}
   {
        $\theta^0 \leftarrow \theta_{t-1}$
        
   		\For{k = 1 to K}
   		{
   		    
       		$\mathbf{X} \sim \{\mathbf{X}\}$\tcp{image batch sample}
       		
       		$\sigma_n \sim rand(0,\sigma_{max})$,~
       		$\mathbf{n} \sim N(0, \sigma_n)$
       		
       		$\mathbf{Y} \leftarrow \mathbf{X} + \mathbf{n}$
       		
       		$\bar{\mathbf{Y}} \leftarrow g(\mathbf{Y})$
       		\tcp*{$g$:non-trainable}

       	    $\sigma_r \sim rand(0,\sigma_{max})$,~
   		    $\mathbf{r} \sim N(0, \sigma_r)$ 
   		
   		    $\mathbf{Z} \leftarrow \bar{\mathbf{Y}} + \mathbf{r}$
   		    
   		    $\theta^k \leftarrow \argminA_{\theta} Loss(\theta|\bar{\mathbf{Y}},\mathbf{Z},\theta^{k-1})$
   		    \tcp*{loss in (\ref{eq_loss_proposed2}) with ADAM}
   		}
   		$\theta_t \leftarrow \theta_{t-1} + \epsilon ( \theta^K - \theta_{t-1} )\newline$
   		\tcp*{$\epsilon$: small update step}
   }
\caption{\newline 
Training via meta-learning. (Stage 0)}
\label{algorithm_reptile}
\end{algorithm}

\begin{algorithm}[h]
\DontPrintSemicolon
  \KwInput{Noisy input: $\mathbf{Y}$, Initial params.: $\theta_T$, $N$}
  \KwOutput{Denoised image: $\tilde{\mathbf{X}}\newline$}

  $\bar{\mathbf{Y}} \leftarrow g(\mathbf{Y})$

  \For{n = 1 to N}
   {
        $\sigma =
        \begin{cases}
           Const. & \text{if known (non-blind)}\\
          rand(0,\sigma_{max}) & \text{otherwise (blind)}
        \end{cases}$
   		
   		$\mathbf{r} \sim N(0, \sigma)$
   		
   		$\mathbf{Z} \leftarrow \bar{\mathbf{Y}} + \mathbf{r}$
   		
   		$\theta' \leftarrow \argminA_{\theta} Loss(\theta|\bar{\mathbf{Y}},\mathbf{Z},\theta_{T+n-1})$
   		\tcp*{loss in (\ref{eq_loss_proposed2}) with ADAM}
   		
   	    $\theta_{T+n} \leftarrow \theta_{T+n-1} + \epsilon ( \theta' - \theta_{T+n-1})\newline$
   	    \tcp*{$\epsilon$: small update step}
   }
   
   $\tilde{\mathbf{X}} \leftarrow f(\mathbf{Y};\theta_{T+N})$

\caption{\newline 
Inference through adaptation. (Stage 1 + Stage 2)}
\label{algorithm_reptile_inference}
\end{algorithm}


\section{Experiments}
Please refer to our supplementary material for the extensive experimental results, and the code will be publicly available upon acceptance.

\subsection{Implementation details}
In our experiments, we evaluate the proposed methods using different state-of-the-art denoisers on DIV2K, Urban100, and BSD68 datasets.
We first pre-train the state-of-the-art denoisers under fair conditions using an NVIDIA 2080Ti graphics card. DnCNN~\cite{DnCNN}, RIDNet~\cite{RIDNet}, and RDN~\cite{RDN} are trained on the DIV2K training set with Gaussian noise until convergence.

Currently, RDN shows the best performance on public benchmark
tests~\cite{benchmark} in removing Gaussian noise, and recent RIDNet shows competitive results. We use the light version of RDN (D = 10,C = 4,G = 16) due to the limited memory size of our graphics unit. 
The standard deviation of the Gaussian noise is randomly selected from [0, 50] during pre-training, and a conventional data augmentation technique is applied. We minimize the distance between the ground truth image and the prediction. 

For meta-learning, we set T = 2000, K = 256, $\sigma_{max}$ = 50, and $\epsilon$ = 1e-5 in Algorithm~\ref{algorithm_reptile} and in Algorithm~\ref{algorithm_reptile_inference},
and the pre-trained networks (i.e., DnCNN, RIDNet, RDN) are used as denoiser $g$ in Fig.~\ref{fig_overall}.

\subsection{Self-similarity exploitation}

\begin{figure}[h]
\begin{center}
\includegraphics[width=1.0\linewidth]{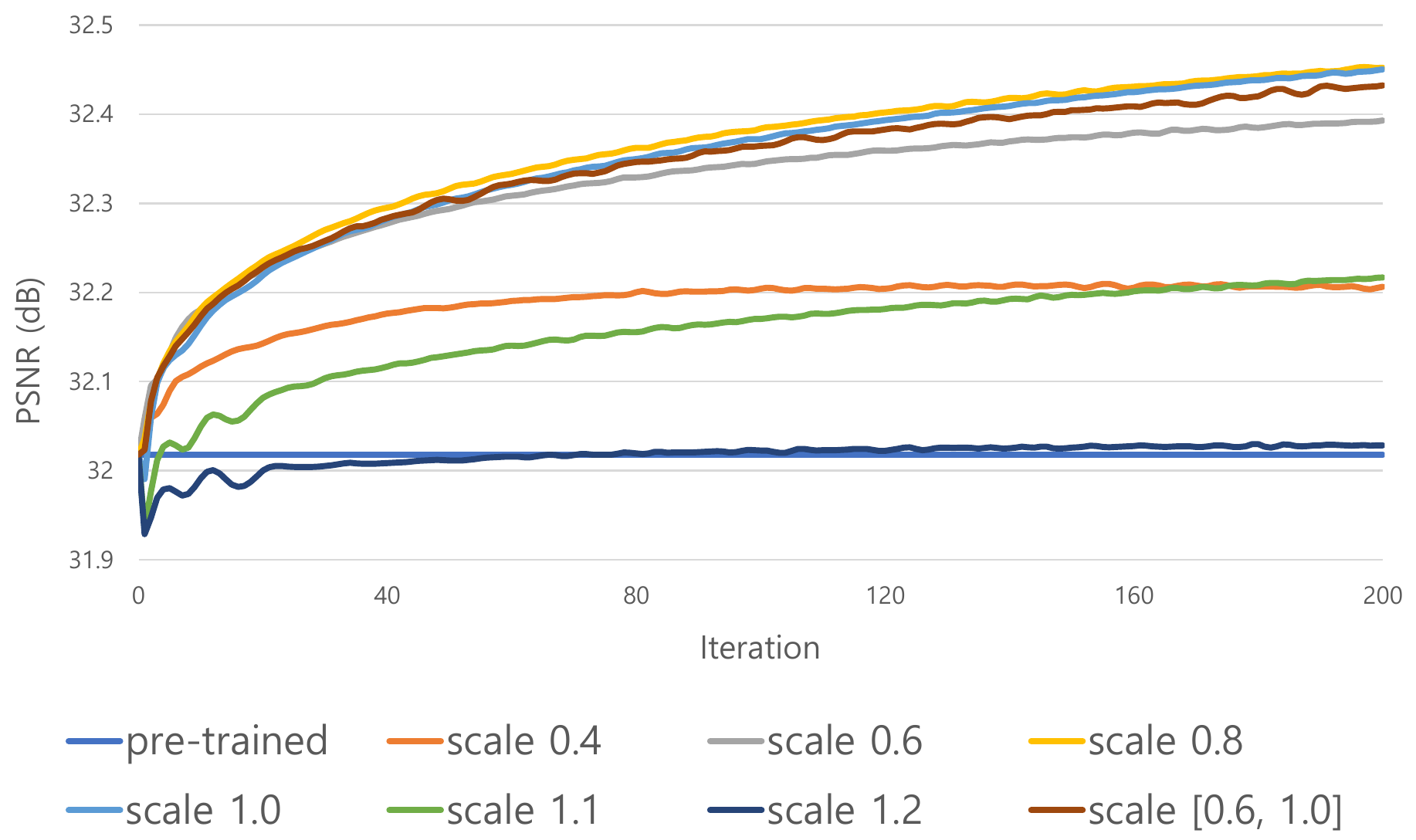}
\end{center}
\caption{Denoising results with our two-phase denoising algorithm.
Performances are evaluated by changing the image scales used in update. DnCNN is used for removing a Gaussian noise ($\sigma=20$) on the Urban100 dataset.}
\label{DnCNN_self-supervision}

\begin{center}
\includegraphics[width=1.0\linewidth]{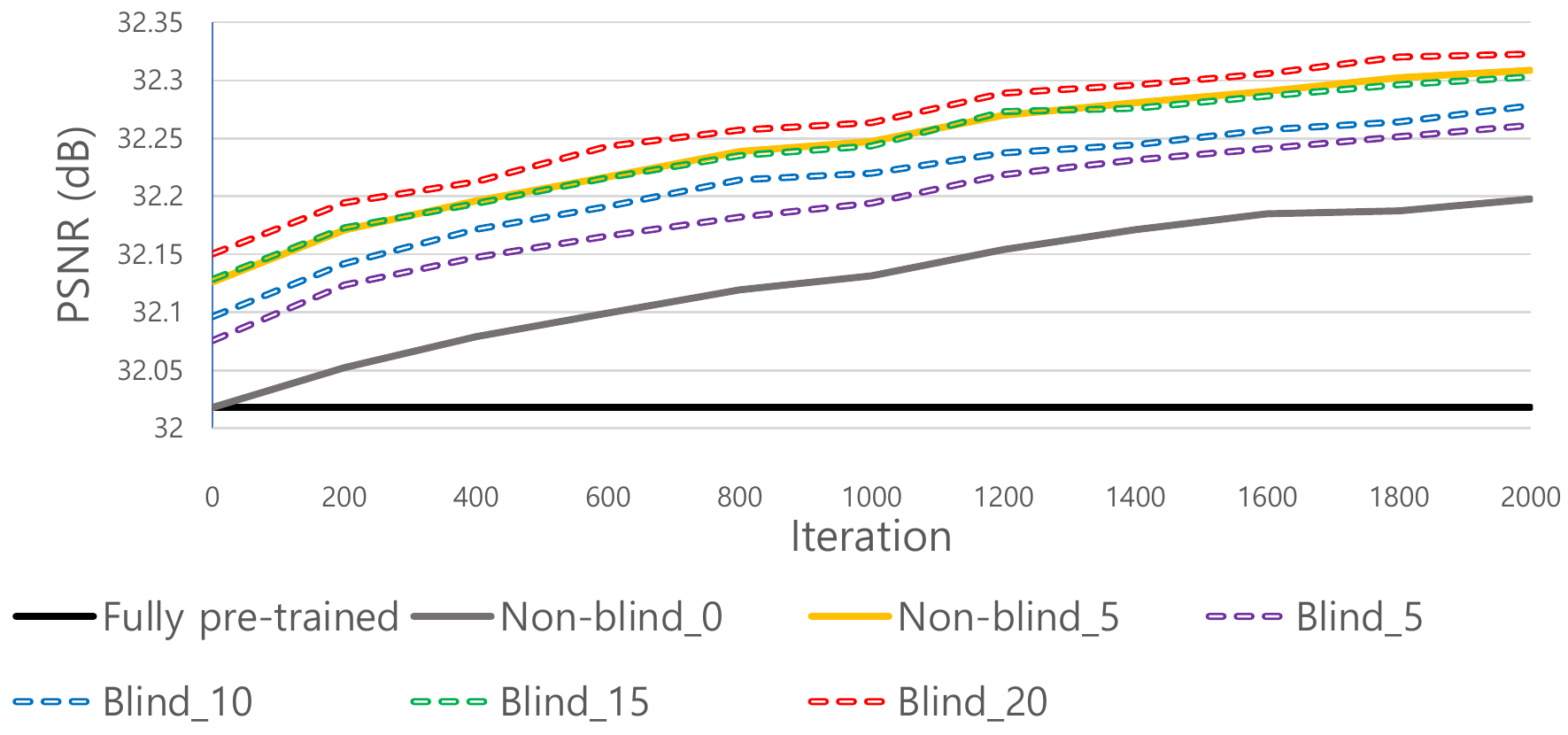}
\end{center}
\caption{Performance evaluation during meta-learning procedure for blind and non-blind denoising with DnCNN.}
\label{fig_meta_learning_urban}
\end{figure}

First, we fine-tune the fully trained DnCNN for 200 iterations on the Urban100 dataset using the proposed two-phase denoising algorithm (w/o meta-learning). 
For the updates, we use different image scales and resize $\bar{\mathbf{Y}}$ with different scaling factors from 0.4 to 1.2.
At each update, we measure the average PSNR values by removing Gaussian noise with $\sigma = 20$. The results are shown in Fig.~\ref{DnCNN_self-supervision}.
As we expected, PSNR values still increase as $N$ increases at different image scales because similar patches are existing across different image scales, as shown in Fig.~\ref{fig_self_sim}. 
Interestingly, we can achieve the best performance when the scaling factor is 0.8 and $N <200$ because the noise level (i.e., residual noise $Var(\mathbf{n}'$)) further decreases by resizing.
Moreover, we also perform updates by choosing the scale randomly using a normal distribution $\sim$ $\mathcal{N}$($\mu$ = 0.8, $\sigma$ = 0.1). The chosen scale value is clipped if it is larger than 1.0 or smaller than 0.6. Thus, we can update the networks using multiples scales (solid brown line). Although, the final performance of the multi-scale training after 200 iterations is lower compared to the result obtained when the scale factor is 0.8 (solid yellow line), we use a random multiple-scale factor in Algorithm~\ref{algorithm_reptile_inference} during inference because the multi-scale version takes less time with smaller images and shows slightly better performances when the number of iterations is small ($\sim$ 5), which well suits to our real testing scenario. During the meta-learning procedure, we use a fixed scale factor $(= 1)$ in Algorithm~\ref{algorithm_reptile}.

We also perform updates for 200 iterations on the BSD68, and DIV2K test sets as well as the Urban100 dataset under the same condition (scale factor = 0.8). After 200 iterations, we measure the performance gain, and the results are given in Table.~\ref{table_psnr_gain}. The gain from the Urban100 dataset is much larger than others because urban images generally include a large number of self-similar patches from man-made repeated structures.

We perform additional experiments to see whether self-similarity is a significant factor in the proposed method. We fine-tune the fully pre-trained DnCNN on the Urban100 dataset, and update the parameters with different sizes of patches. To be specific, we collect 64$\times$64 and 128$\times$128 patches from the Urban100 dataset where the 64$\times$64 patches are centrally cropped version of the randomly chosen 128$\times$128 patches. We compare the PSNR values obtained results by parameters learned from 128$\times$128 patches and 64$\times$64 patches respectively. For the evaluation of parameters updated with 128$\times$128 patches, we measure the PSNR on the 64$\times$64 central parts of the 128$\times$128 patches to carry out a fair comparison. The comparison results on different Gaussian noises are given in Table.~\ref{table_self_similarity}, and the parameters updated with larger patches render better results because larger patches are likely to include more corresponding patches (i.e., large $M$).

In this ablation study, we demonstrate that our algorithm can exploit the self-similarity within the given input, and thus the proposed method can produce better results where $N$ and $M$ are large.

\begin{table}[]
\centering
\begin{tabular}{lllll}
\cline{1-4}
\multicolumn{1}{|l|}{} & \multicolumn{1}{l|}{Urban100} & \multicolumn{1}{l|}{BSD68} & \multicolumn{1}{l|}{DIV2K} &  \\ \cline{1-4}
\multicolumn{1}{|l|}{PSNR gain} & \multicolumn{1}{c|}{0.44}  & \multicolumn{1}{c|}{0.17}  & \multicolumn{1}{c|}{0.23}  &  \\ \cline{1-4}&&&&  \\
& &&& 
\end{tabular}
\vspace{-3Ex}
\caption{After 200 updates, PSNR gains on Urban100, BSD68, and DIV2K test set are measured. Performance gain is particularly huge on the Urban100 dataset.}
\label{table_psnr_gain}
\vspace{1Ex}
\end{table}


\begin{table}[]
\centering
\begin{tabular}{lllll}
\hline
\multicolumn{1}{|l|}{}      
& \multicolumn{1}{l|}{$\sigma$ = 10} 
& \multicolumn{1}{l|}{$\sigma$ = 20}    
& \multicolumn{1}{l|}{$\sigma$ = 30}    
& \multicolumn{1}{l|}{$\sigma$ = 40}    \\ \hline
\multicolumn{1}{|l|}{Small patch} & \multicolumn{1}{c|}{35.94}                    & \multicolumn{1}{l|}{32.52} & \multicolumn{1}{l|}{30.68} & \multicolumn{1}{c|}{29.15} \\ \hline
\multicolumn{1}{|l|}{Large patch} & \multicolumn{1}{c|}{35.94}                    & \multicolumn{1}{l|}{32.57} & \multicolumn{1}{l|}{30.74} & \multicolumn{1}{c|}{29.23} \\ 
\hline&&&&                           
\end{tabular}
\caption{Performance comparison by updating DnCNN using different sizes of patches. Parameters trained with large patches provide consistently better results for various Gaussian noise levels.}
\label{table_self_similarity}
\end{table}


\subsection{Denoising results via meta-learning}

In Fig.~\ref{fig_meta_learning_urban}, we evaluate the performance of DnCNN during the meta-learning procedure in Algorithm~\ref{algorithm_reptile}. We use DIV2K training set for meta-learning and set $T=2000$. At each meta-learning iteration, degraded Urban100 dataset with Gaussian noise ($\sigma=20$) is restored using the method in Algorithm~\ref{algorithm_reptile_inference}. We measure the performance by changing the number of iterations $N$ in Algorithm~\ref{algorithm_reptile_inference} from 0 to 20 in blind and non-blind manner.
As meta-learning progresses (i.e., $t\rightarrow T)$ inference accuracy improves gradually. DnCNN provides consistently superior results with large $N$ for both blind and non-blind denoising.

In Table~\ref{table_meta_results}, we provide quantitative comparisons results. Blind and non-blind denoising results from meta-learned RIDNet (MetaRIDNet), RDN (MetaRDN), and DnCNN (MetaDnCNN) are measured with different settings, and compared with results by conventional methods (BM3D~\cite{BM3D}, MemNet~\cite{memnet}, and FFDNet~\cite{FFDNet}). Our meta-learned blind/non-blind denoisers can produce better results with a small number of updates because they can adapt their parameters quickly to the specific input, and can outperform the pre-trained baseline models with only 5 iterations. Note that the performance gaps between the naive fine-tuning (Finetune$\_$5) and our meta-learning-based adaptation (Bind$\_$5) for 5 iterations are large particularly when the noise level is high, and these results demonstrate that the proposed method can improve the performance more quickly than naive fine-tuning. With NVIDIA 2080Ti Graphics card, it takes around 0.99, 2.49, and 2.64 seconds to restore a 1000$\times$600 image with MetaDnCNN, MetaRDN, and MetaRIDNet updated for 5 times respectively.

\begin{figure*}[h]
\begin{center}
\includegraphics[width=1.0\linewidth]{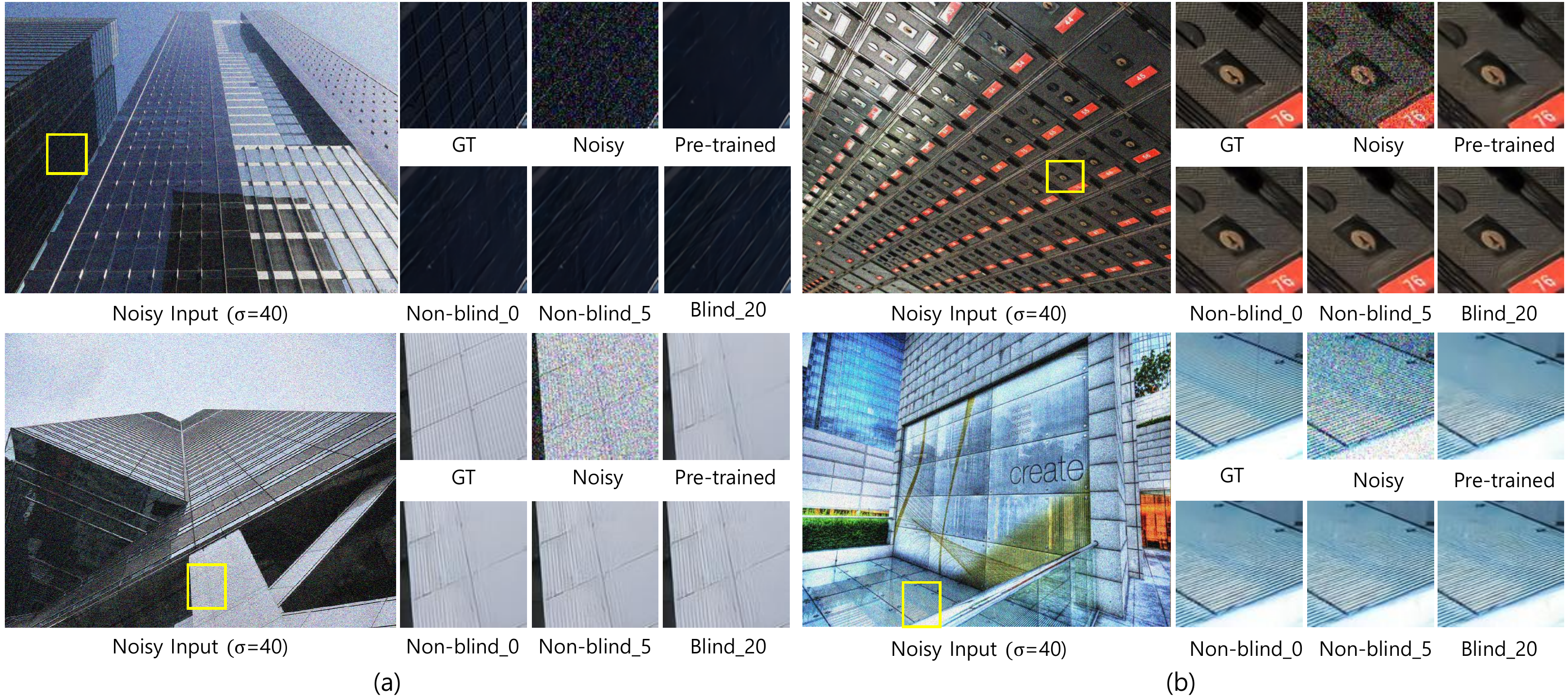}
\end{center}
\caption{Visual comparisons. (a) Denoising results with our MetaRIDNet~\cite{RIDNet}. (b) Denoising results with our MetaRDN~\cite{RDN}. Notably, Non-blind$\_0$ denotes the results obtained by parameters $\theta_T$ which is the initial parameters of the inference step in Algorithm~\ref{algorithm_reptile_inference}.}
\label{fig_visual_comparison}
\vspace{1Ex}
\end{figure*}

In Fig.~\ref{fig_visual_comparison}, we provide qualitative comparison results. The inputs are corrupted with high-level Gaussian noise ($\sigma$ = 40), and the proposed methods restore the clean images in blind and non-blind manners. In particular, with more iterations during inference, our blind and non-blind methods can produce visually much better results and restore tiny details compared to the fully pre-trained baseline models.

\begin{table*}[h]
\centering
\resizebox{\textwidth}{!}{
\begin{tabular}{|c|c|c|c|c|c|c|c|c|c|c|c|c|c|}
\hline
       & Dataset                                                         & \multicolumn{4}{c|}{Urban100}
       & \multicolumn{4}{c|}{DIV2K}
       & \multicolumn{4}{c|}{BSD68}
       \\ \hline
Method & Adaptation\backslashbox{}{Noise}                                          
& $\sigma$ = 10 & $\sigma$ = 20 & $\sigma$ = 30 & $\sigma$ = 40
& $\sigma$ = 10 & $\sigma$ = 20 & $\sigma$ = 30 & $\sigma$ = 40
& $\sigma$ = 10 & $\sigma$ = 20 & $\sigma$ = 30 & $\sigma$ = 40
\\ \hline

BM3D~\cite{BM3D} & - & 35.77 & 31.92 & 29.38 & 27.06 & 36.15 & 32.13 & 29.58 & 27.49 & 35.75 & 31.52 & 29.07 & 27.19 \\ \hline
MemNet~\cite{memnet} & - & 35.66 & 32.32 & 30.32 & 28.87 & 36.60 & 33.01 & 30.97 & 29.55 & 36.07 & 32.27 & 30.21 & 28.83 \\ \hline
FFDNet~\cite{FFDNet} & - & 35.43 & 31.87 & 29.51 & 27.60 & 36.36 & 32.55 & 30.08 & 28.11 & 35.82 & 31.87 & 29.55 & 27.82 \\ \hline

\begin{tabular}[c]{@{}c@{}}MetaRIDNet\\ (ours)\end{tabular} & \begin{tabular}[c]{@{}c@{}}Fully pre-trained\\ Finetune\_5\\ Non-blind\_5\\ Blind\_5\\ Blind\_10\\ Blind\_15\\ Blind\_20\end{tabular} &
\begin{tabular}[c]{@{}c@{}}35.67\\ 35.76\\ \color{Red}35.94\\ 35.88\\ 35.86\\ 35.86\\ 35.89\end{tabular} & 
\begin{tabular}[c]{@{}c@{}}32.40\\ 32.51\\ 32.74\\ 32.70\\ 32.72\\ 32.75\\ \color{Red}32.76\end{tabular} & 
\begin{tabular}[c]{@{}c@{}}30.44\\ 30.56\\ 30.83\\ 30.79\\ 30.83\\ 30.85\\ \color{Red}30.87\end{tabular} & 
\begin{tabular}[c]{@{}c@{}}29.02\\ 29.14\\ 29.43\\ 29.39\\ 29.43\\ 29.46\\ \color{Red}29.48\end{tabular} & 
\begin{tabular}[c]{@{}c@{}}36.63\\ 36.68\\ 36.83\\ 36.74\\ 36.77\\ 36.78\\ \color{Red}36.79\end{tabular} & 
\begin{tabular}[c]{@{}c@{}}33.08\\ 33.13\\ 33.28\\ 33.24\\ 33.26\\ 33.28\\ \color{Red}33.29\end{tabular} & 
\begin{tabular}[c]{@{}c@{}}31.06\\ 31.12\\ 31.28\\ 31.24\\ 31.27\\ 31.28\\ \color{Red}31.30\end{tabular} & 
\begin{tabular}[c]{@{}c@{}}29.65\\ 29.72\\ 29.88\\ 29.84\\ 29.87\\ 29.89\\ \color{Red}29.90\end{tabular} & 
\begin{tabular}[c]{@{}c@{}}36.10\\ 36.15\\ \color{Red}36.22\\ 36.17\\ 36.18\\ 36.19\\ 36.20\end{tabular} & 
\begin{tabular}[c]{@{}c@{}}32.32\\ 32.37\\ 32.46\\ 32.44\\ 32.44\\ 32.45\\ \color{Red}32.47\end{tabular} & 
\begin{tabular}[c]{@{}c@{}}30.27\\ 30.33\\ 30.42\\ 30.40\\ 30.42\\ 30.43\\ \color{Red}30.44\end{tabular} & 
\begin{tabular}[c]{@{}c@{}}28.89\\ 28.96\\ 29.06\\ 29.03\\ 29.05\\ 29.07\\ \color{Red}29.07\end{tabular} \\ \hline

\begin{tabular}[c]{@{}c@{}}MetaRDN\\ (ours)\end{tabular} & \begin{tabular}[c]{@{}c@{}}Fully pre-trained\\ Finetune\_5\\ Non-blind\_5\\ Blind\_5\\ Blind\_10\\ Blind\_15\\ Blind\_20\end{tabular} &
\begin{tabular}[c]{@{}c@{}}35.46\\ 35.55\\ 35.76\\ 35.71\\ 35.72\\ 35.73\\ 35.74\end{tabular} & \begin{tabular}[c]{@{}c@{}}32.11\\ 32.22\\ 32.50\\ 32.47\\ 32.51\\ 32.53\\ 32.55\end{tabular} & \begin{tabular}[c]{@{}c@{}}30.10\\ 30.22\\ 30.55\\ 30.52\\ 30.57\\ 30.60\\ 30.63\end{tabular} & \begin{tabular}[c]{@{}c@{}}28.65\\ 28.77\\ 29.12\\ 29.08\\ 29.14\\ 29.17\\ 29.20\end{tabular} & \begin{tabular}[c]{@{}c@{}}36.44\\ 36.53\\ 30.69\\ 36.64\\ 36.64\\ 36.66\\ 36.67\end{tabular} & \begin{tabular}[c]{@{}c@{}}32.87\\ 32.94\\ 33.13\\ 33.11\\ 33.13\\ 33.14\\ 33.16\end{tabular} & \begin{tabular}[c]{@{}c@{}}30.82\\ 30.90\\ 31.09\\ 31.08\\ 31.10\\ 31.12\\ 31.13\end{tabular} & \begin{tabular}[c]{@{}c@{}}29.40\\ 29.48\\ 29.68\\ 29.65\\ 29.68\\ 29.69\\ 29.71\end{tabular} & \begin{tabular}[c]{@{}c@{}}35.99\\ 36.07\\ 36.14\\ 36.11\\ 36.11\\ 36.11\\ 36.13\end{tabular} & \begin{tabular}[c]{@{}c@{}}32.19\\ 32.26\\ 32.36\\ 32.35\\ 32.37\\ 32.37\\ 32.39\end{tabular} & \begin{tabular}[c]{@{}c@{}}30.12\\ 30.18\\ 30.31\\ 30.30\\ 30.32\\ 30.33\\ 30.34\end{tabular} & \begin{tabular}[c]{@{}c@{}}28.74\\ 28.80\\ 28.93\\ 28.91\\ 28.94\\ 28.95\\ 28.96\end{tabular} \\ \hline

\begin{tabular}[c]{@{}c@{}}MetaDnCNN\\ (ours)\end{tabular} & \begin{tabular}[c]{@{}c@{}}Fully pre-trained\\ Finetune\_5\\ Non-blind\_5\\ Blind\_5\\ Blind\_10\\ Blind\_15\\ Blind\_20\end{tabular} &
\begin{tabular}[c]{@{}c@{}}35.46\\ 35.55\\ 35.68\\ 35.57\\ 35.56\\ 35.57\\ 35.59\end{tabular} & \begin{tabular}[c]{@{}c@{}}32.01\\ 32.12\\ 32.30\\ 32.25\\ 32.28\\ 32.30\\ 32.32\end{tabular} & \begin{tabular}[c]{@{}c@{}}29.97\\ 30.09\\ 30.30\\ 30.26\\ 30.29\\ 30.31\\ 30.35\end{tabular} & \begin{tabular}[c]{@{}c@{}}28.48\\ 28.62\\ 28.85\\ 28.79\\ 28.81\\ 28.84\\ 28.87\end{tabular} & \begin{tabular}[c]{@{}c@{}}36.41\\ 36.46\\ 36.56\\ 36.47\\ 36.49\\ 36.51\\ 36.52\end{tabular} & \begin{tabular}[c]{@{}c@{}}32.80\\ 32.85\\ 32.98\\ 32.93\\ 32.94\\ 32.96\\ 32.97\end{tabular} & \begin{tabular}[c]{@{}c@{}}30.76\\ 30.81\\ 30.94\\ 30.87\\ 30.91\\ 30.93\\ 30.94\end{tabular} & \begin{tabular}[c]{@{}c@{}}29.34\\ 29.39\\ 29.51\\ 29.43\\ 29.47\\ 29.48\\ 29.51\end{tabular} & \begin{tabular}[c]{@{}c@{}}35.99\\ 36.04\\ 36.09\\ 36.02\\ 36.03\\ 36.04\\ 36.05\end{tabular} & \begin{tabular}[c]{@{}c@{}}32.16\\ 32.21\\ 32.29\\ 32.25\\ 32.26\\ 32.28\\ 32.28\end{tabular} & \begin{tabular}[c]{@{}c@{}}30.09\\ 30.14\\ 30.22\\ 30.17\\ 30.19\\ 30.21\\ 30.22\end{tabular} & \begin{tabular}[c]{@{}c@{}}28.70\\ 28.73\\ 28.82\\ 28.79\\ 28.80\\ 28.81\\ 28.83\end{tabular} \\ \hline

\end{tabular}
}
\vspace{1Ex}
\caption{
Quantitative comparisons.
Non-blind$\_N$ and Blind$\_N$ indiciate that the network is updated for $N$ iterations during the testing phase with and without knowing the Gaussian noise level (non-blind and blind denoising respectively). 
}
\label{table_meta_results}
\end{table*}

\section{Conclusion}
Considering that we can improve the performance of the conventional supervision-based denoising methods during test time using the self-similarity property from the given noisy input image with the proposed loss function. Thus, we introduce a new two-phase denoising approach that allows the update of the network parameters from the fully trained version at test time and enhance the image quality significantly by exploiting self-similarity. Furthermore, we integrate meta-learning technique while updating (fine-tuning) our denoiser to enable quick parameter adaptation and accurate inference at test time.
Our proposed algorithm can be generally applicable to many denoising networks, and we improve the restoration quality significantly without changing the architectures of the state-of-the-art denoising methods. Experimental results demonstrate the superiority of the proposed method.

\section*{Acknowledgement}
This work was supported by the research fund of SK Telecom T-Brain
and Hanyang University(HY-2018).

\newpage
{\small
\bibliographystyle{ieee_fullname}
\bibliography{denoising}
}

\end{document}